\documentclass[letterpaper, 10 pt, conference]{ieeeconf}  % Comment this line out if you need a4paper

\IEEEoverridecommandlockouts                              % This command is only needed if
\usepackage{graphics} % for pdf, bitmapped graphics files
\usepackage{epsfig} % for postscript graphics files
\usepackage{times} % assumes new font selection scheme installed
\usepackage{amsmath} % assumes amsmath package installed
\usepackage{amssymb}  % assumes amsmath package installed
\usepackage{makecell}
\usepackage{booktabs}
\usepackage{multirow}
\usepackage{multicol}
\usepackage{here}
\usepackage{kotex}
\usepackage{array}
\usepackage[table]{xcolor} % loads colortbl; MUST be before arydshln, else \cellcolor breaks
\usepackage{arydshln}      % dashed table rules (\cdashline); load after colortbl
\usepackage{caption}
\usepackage{cuted}         % for the full-width teaser strip on page 1
\usepackage{threeparttable}
\usepackage{graphicx}
\graphicspath{{figs/}{./}}   % look for figures in figs/ first
\usepackage{stfloats}
\usepackage{hyperref}
\usepackage{cleveref}      % for \cref / \Cref
\crefname{table}{Table}{Tables}
\crefname{figure}{Fig.}{Figs.}
\crefname{section}{Section}{Sections}
\crefname{subsection}{Section}{Sections}
\crefname{subsubsection}{Section}{Sections}
\crefname{equation}{Eq.}{Eqs.}

\definecolor{bestcolor}{HTML}{bce6cd}
\definecolor{secondcolor}{HTML}{e4eebc}
\definecolor{thirdcolor}{HTML}{fef8c4}

\title{\LARGE \bf
LEGO-SLAM: Language-Embedded Gaussian Optimization SLAM
}

\author{Sibaek Lee, Seongbo Ha, Kyeongsu Kang, Joonyeol Choi, Seungjun Tak, and Hyeonwoo Yu%
\thanks{This work has been submitted to the IEEE for possible publication. Copyright may be transferred without notice, after which this version may no longer be accessible. The authors are with the Department of Intelligent Robotics, Sungkyunkwan University, Suwon, South Korea. {\tt\small \{lmjlss, sobo3607, thithin0821, joonyeol99, tmdwns8840, hwyu\}@g.skku.edu}}%
}

\begin{document}

% Full-width teaser placed inside the title area (avoids the cuted strip + \thanks footnote clash).
\IEEEaftertitletext{%
\begin{center}
\vspace{-0.5\baselineskip}% pull the teaser up toward the author line
\includegraphics[width=\textwidth]{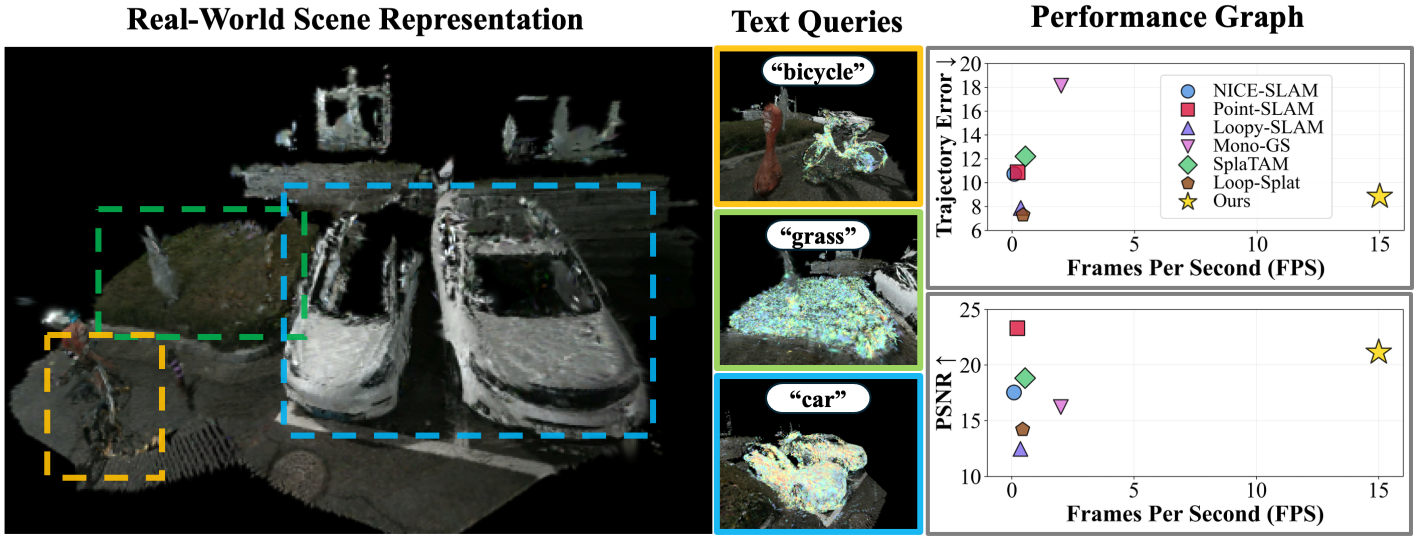}
\captionof{figure}{LEGO-SLAM: real-time, open-vocabulary 3DGS-SLAM. (Left) A large-scale, self-captured real-world outdoor scene reconstructed online by our system, with colored boxes highlighting specific regions. (Middle) Relevancy maps showing the 3D localization for corresponding text queries. (Right) Graphs on the ScanNet dataset \cite{dai2017scannet} demonstrate LEGO-SLAM operates at 15 FPS while maintaining competitive performance.}
\label{fig:title_figure}
% \vspace{-0.5\baselineskip}% let the two-column text start a bit higher
\end{center}
}

\maketitle
\thispagestyle{empty}
\pagestyle{empty}

%%%%%%%%%%%%%%%%%%%%%%%%%%%%%%%%%%%%%%%%%%%%%%%%%%%%%%%%%%%%%%%%%%%%%%%%%%%%%%%%
\begin{abstract}
  Recent advances in 3D Gaussian Splatting (3DGS) have enabled Simultaneous Localization and Mapping (SLAM) systems to build photorealistic maps. However, these maps lack the open-vocabulary semantic understanding required for robotic interaction. Integrating language features into SLAM remains a significant challenge, as storing high-dimensional features incurs excessive memory and rendering overhead, while existing methods with static models lack adaptability for novel environments. We propose LEGO-SLAM (Language-Embedded Gaussian Optimization SLAM), a framework that achieves real-time, open-vocabulary mapping within a 3DGS-based SLAM system. At the core of our method is a scene-adaptive autoencoder that distills high-dimensional language embeddings into a compact 16-dimensional feature space, reducing the memory per Gaussian and accelerating rendering. Unlike static approaches, our encoder adapts online to unseen scenes. These compact features also enable a language-guided pruning strategy that identifies semantic redundancy, reducing the map's Gaussian count by up to 58\% while maintaining rendering quality. Furthermore, we introduce a language-based loop detection approach that reuses the language features already extracted for mapping, eliminating the need for a separate detection model. Experiments demonstrate that LEGO-SLAM achieves competitive mapping quality and tracking accuracy, all while providing open-vocabulary capabilities at 15 FPS. Our project page is available at \url{https://lab-of-ai-and-robotics.github.io/LEGO-SLAM/}.
\end{abstract}

\section{Introduction}
\label{sec:intro}
While the foundational objective of Simultaneous Localization and Mapping (SLAM) has long been to jointly construct a map and estimate an agent’s location, technological advancements have elevated the ambition towards generating rich and scalable environmental representations. In response, recent breakthroughs in 3D reconstruction have enhanced the mapping component of SLAM, with Neural Radiance Fields (NeRF) \cite{mildenhall2021nerf} and 3D Gaussian Splatting (3DGS) \cite{kerbl20233d} emerging as prominent solutions for creating photorealistic maps \cite{tosi2024nerfs}. NeRF provided memory-efficient implicit mapping but failed to meet the real-time demands of SLAM due to slow rendering. This limitation was overcome by 3DGS, which delivers high-fidelity mapping in real time.
However, such photorealistic maps lack the semantic understanding required for enabling embodied AI to perform diverse downstream tasks \cite{shafiullah2022clip, lee2025lamp, shah2023lm}. The evolution of these methods has progressed beyond simple RGB representation to semantic mapping, assigning categorical labels to the reconstructed geometry \cite{li2024sgs, zhu2024sni, li2024dns, rosinol2020kimera, zhu2024semgauss}. Yet, these approaches have relied on a closed-set paradigm, limiting them to predefined object labels. While the 3D reconstruction field is rapidly advancing towards open-vocabulary representations that can interpret arbitrary language queries \cite{kerr2023lerf, liao2024ov, qin2024langsplat, zhou2024feature}, integrating these capabilities into SLAM remains a challenge, due to critical limitations in real-time performance and model adaptability.

Open-vocabulary 3D reconstruction approaches that store high-dimensional language features on each Gaussian suffer from rendering latency and high memory overhead, making them impractical for real-time SLAM \cite{kerr2023lerf, liao2024ov, zhou2024feature}. Furthermore, methods that rely on pretrained models to extract fixed, low-dimensional features lack the adaptability required for SLAM systems, which must continuously learn from new environments \cite{qin2024langsplat}. More recently, an online approach~\cite{katragadda2025online} also targets real-time open-vocabulary Gaussian mapping, but it falls short of real-time operation, leaving fast online open-vocabulary SLAM an open problem. To address this, we propose LEGO-SLAM (Language-Embedded Gaussian Optimization SLAM), a framework that achieves real-time, open-vocabulary mapping within a 3DGS-based SLAM system.
Our core contribution is a scene-adaptive autoencoder that distills high-dimensional language embeddings into a compact 16-dimensional space, trained online during mapping. Rather than decoding the whole map back to a high dimension for querying, we project the text query into this compact space for fast comparison. The choice of 16 dimensions provides the best balance between memory per Gaussian and rendering speed, as validated in \cref{sec:ablation}. Because such abstract features cannot be initialized directly from RGB-D, we leverage a pretrained encoder as a prior to ensure the rapid convergence essential for real-time SLAM.

Beyond mapping, these compact features benefit other SLAM components. They enable a language-guided pruning strategy that identifies semantically redundant Gaussians with negligible overhead, addressing the trade-offs of geometric pruning \cite{kerbl20233d, keetha2024splatam, matsuki2024gaussian, ha2024rgbd, zhu2025loopsplat} without sacrificing detail. The language features already extracted as distillation targets are further reused for loop detection, performing place recognition without a separate detection model \cite{arandjelovic2016netvlad, mur2017orb}, which corrects long-term drift while keeping tracking accuracy competitive with leading systems \cite{liso2024loopy, zhu2025loopsplat}. We demonstrate the effectiveness of our framework on standard benchmarks including Replica \cite{straub2019replica}, TUM-RGBD \cite{sturm2012benchmark} and ScanNet \cite{dai2017scannet}, and our main contributions are as follows:

\begin{itemize}
    \item We propose LEGO-SLAM, a real-time, open-vocabulary 3DGS SLAM framework that performs high-fidelity mapping and tracking simultaneously.

    \item A lightweight, scene-adaptive autoencoder that learns a compact 16-dimensional language representation with a pretrained prior for rapid convergence, enabling real-time, low-overhead mapping and fast open-vocabulary queries.

    \item A language-guided Gaussian pruning strategy that improves memory efficiency by removing redundant Gaussians without sacrificing structural detail.

    \item An efficient language-based loop detection approach that reuses the features extracted for mapping, avoiding a separate model while ensuring robust long-term tracking accuracy.
\end{itemize}

\section{Related Work}
\label{sec:Related_work}
\subsection{Neural Rendering SLAM}
Neural Rendering SLAM moves beyond traditional geometric representations to build high-fidelity, photorealistic maps by optimizing a scene representation against input images \cite{tosi2024nerfs}.
Early works in this domain integrated NeRF \cite{mildenhall2021nerf} into the SLAM pipeline, demonstrating the ability to construct continuous implicit map representations from RGB images \cite{zhu2022nice, sandstrom2023point, liso2024loopy}. While they produced maps of high visual quality, the computational cost of per-ray rendering made them unsuitable for live SLAM operations. 3DGS \cite{kerbl20233d} addressed this bottleneck with an explicit representation that enables real-time rendering, leading to its rapid adoption in SLAM \cite{keetha2024splatam, yan2024gs, matsuki2024gaussian, yugay2023gaussian}. Efforts have also aimed to improve long-term robustness in large-scale environments by incorporating loop closure into these systems \cite{zhu2025loopsplat}.
However, these photorealistic maps lack the semantic context required for robotic interaction. Existing semantic extensions, including recent Gaussian-based semantic SLAM \cite{li2025hier, li2024gs3lam}, remain within a closed-set paradigm limited to a predefined list of categories \cite{li2024sgs, zhu2024sni, li2024dns, rosinol2020kimera, zhu2024semgauss}.

\subsection{Open-Vocabulary Scene Representation}
The transition to an interactive open-vocabulary map remains a significant hurdle for SLAM.
The open-vocabulary shift was catalyzed by vision-language models like CLIP \cite{radford2021learning}, which established a shared image-text embedding space. Concurrently, foundation models such as DINO \cite{oquab2023dinov2} and SAM \cite{kirillov2023segment} have excelled in tasks like segmentation and tracking \cite{ranftl2021vision, liu2024grounding}, though they often lack native open-vocabulary recognition. Subsequent works have leveraged these to extract dense, per-pixel language features \cite{li2022language}, enabling a new class of open-vocabulary downstream tasks.
These per-pixel descriptors have been lifted into 3D, enabling open-vocabulary 3D reconstruction. However, such methods are not designed for the real-time constraints of SLAM, as distilling high-dimensional features per frame remains a bottleneck \cite{kerr2023lerf, liao2024ov, zhou2024feature, qin2024langsplat}. In contrast, we learn a compact, scene-adaptive language representation directly within a real-time SLAM framework.

\section{Method}
\label{sec:method}
\begin{figure*}[t]
    \centering
    \includegraphics[width=1.0\textwidth]{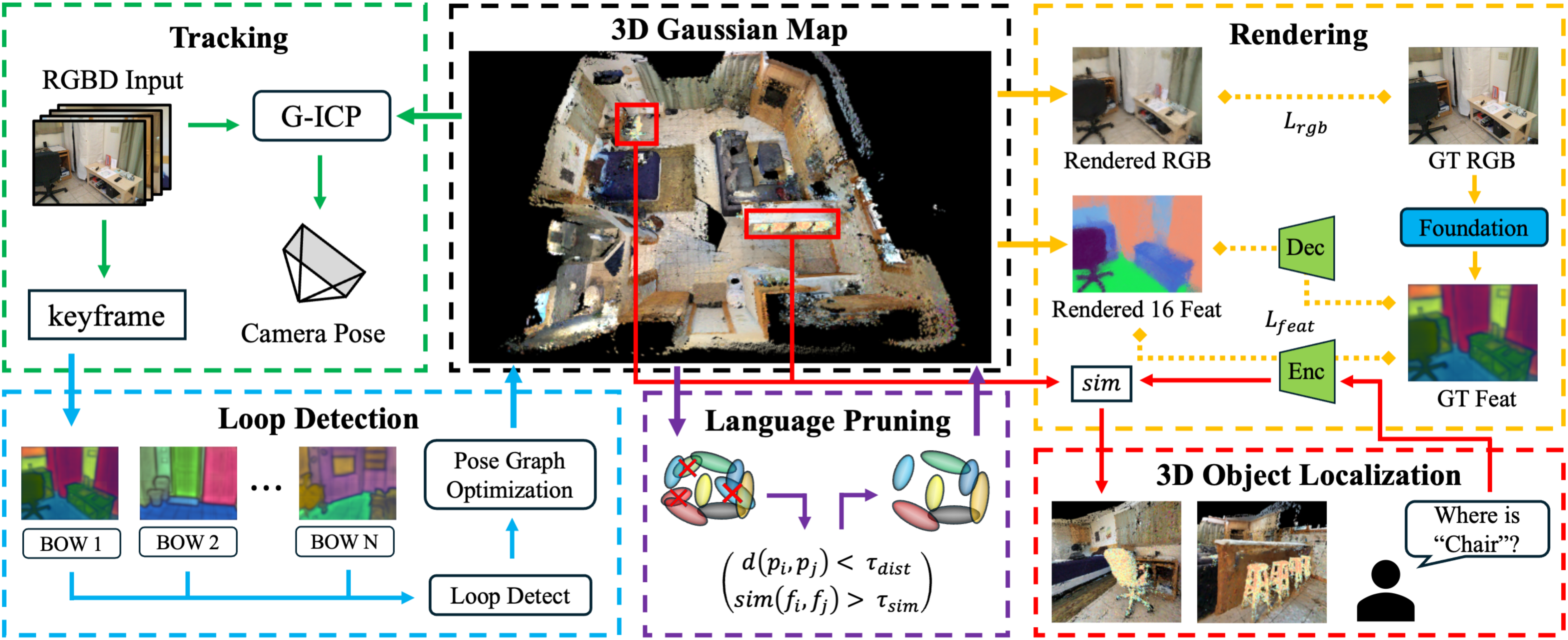}
    \caption{System Overview. LEGO-SLAM architecture, where the Tracking module estimates pose and the Mapping module optimizes the 3D Gaussian Map via language distillation. This map is refined by Language Pruning and Loop Detection, enabling 3D Object Localization.}
    \label{fig:method}
\end{figure*}

LEGO-SLAM constructs a 3D Gaussian Splatting map enriched with open-vocabulary language features in real time. As illustrated in \cref{fig:method}, it couples a Tracking module (\cref{sec:tracking}) with a Mapping module (\cref{sec:mapping}) that optimizes the language-embedded map, and adds a language-guided pruning strategy (\cref{sec:pruning}) and loop detection (\cref{sec:loop_closure}) for efficient, long-term-consistent operation.

\subsection{Tracking}
\label{sec:tracking}
The tracking module estimates the camera pose $T_k \in SE(3)$ for each frame $I_k$ via G-ICP \cite{segal2009generalized} initialized from $T_{k-1}$. Unlike decoupled systems \cite{mur2017orb}, we track directly against our unified 3D Gaussian map, eliminating the need for a separate tracking representation. Source Gaussians generated from the current depth are aligned with target Gaussians sampled from the map. A key efficiency is that source covariances computed during this 3D-to-3D alignment are reused to initialize new Gaussians in the mapping stage, avoiding redundant computation. After pose optimization, a frame is selected as a keyframe if the proportion of inlier correspondences drops below a threshold, ensuring that new Gaussians are added only when significant novel information is observed.

\subsection{Mapping}
\label{sec:mapping}

\subsubsection{Language-Embedded 3D Gaussians}
To enable open-vocabulary capabilities, we extend the standard 3D Gaussian representation. Each Gaussian is defined by a set of optimizable attributes $\Theta = \{ \mathbf{p}, \mathbf{q}, \mathbf{s}, \alpha, \mathbf{c}, \mathbf{f} \}$, which includes position $\mathbf{p}$, rotation $\mathbf{q}$, scale $\mathbf{s}$, opacity $\alpha$, SH color coefficients $\mathbf{c}$, and our compact language feature $\mathbf{f} \in \mathbb{R}^{16}$. Following the standard 3DGS pipeline, these attributes are rendered to produce both an RGB image $\hat{I}$ and a feature map $F_{\text{render}}$.

\subsubsection{Gaussian Initialization}
When a new keyframe is received, we generate new 3D Gaussians from its depth data, deriving their geometric attributes of position, scale, and rotation from the 3D point cloud. To maximize efficiency, we reuse the covariance matrices computed during the G-ICP tracking stage as the initial scale and rotation for these new Gaussians, avoiding redundant calculations.
While geometric attributes are directly initialized, the abstract language features present a unique challenge. To address this, we leverage our scene-adaptive encoder. We first use a pretrained vision-language extractor to obtain a high-dimensional feature map from the keyframe's RGB, which is then passed through our encoder to produce a strong prior for their language features. This encoder-based initialization is critical for the rapid feature convergence essential for real-time SLAM. In addition, before insertion, the keyframe pose is briefly refined by a rendering-based optimization that minimizes an RGB and depth loss over high-confidence pixels, and the refined pose is propagated back to the tracker.

\subsubsection{Map Optimization via Feature Distillation}
Following initialization, we optimize the map over several iterations. In each iteration, we randomly select a keyframe from the active window to render its view and minimize a joint loss function $\mathcal{L}_{\text{total}}$:
\begin{equation}
\label{eq:total_loss}
\mathcal{L}_{\text{total}} = \mathcal{L}_{\text{rgb}} + w_{\text{depth}}\mathcal{L}_{\text{depth}} + w_{\text{feat}}\mathcal{L}_{\text{feat}}
\end{equation}
where $w_{\text{depth}}$ and $w_{\text{feat}}$ are weighting coefficients, $\mathcal{L}_{\text{rgb}}$ is the standard photometric loss combining an L1 and a D-SSIM term between $\hat{I}$ and $I$, and $\mathcal{L}_{\text{depth}}$ is an L1 depth loss between $\hat{D}$ and $D$.
The feature distillation loss $\mathcal{L}_{\text{feat}}$ is the core of our language feature learning:
\begin{equation}
\label{eq:feat_loss} \mathcal{L}_{\text{feat}} = \mathcal{L}_{\text{L1}}(D_{\theta}(F_{\text{render}}), F_{\text{gt}})
\end{equation}
To compute this, we render the feature map $F_{\text{render}}$ from our 3D Gaussians. This map is passed through our lightweight convolutional decoder $D_{\theta}$ to upsample it to the high dimension of the frozen extractor's features $F_{\text{gt}}$ for L1 comparison. This distillation allows our 3D Gaussians to learn a compact and scene-adaptive language representation while maintaining real-time performance. Our framework is agnostic to the vision-language extractor used as the distillation target.

\subsubsection{Scene-Adaptive Encoder Optimization}
We found that jointly optimizing the encoder $E_{\phi}$ with the map introduces training instability, especially if adaptation starts before the Gaussian features $\mathbf{f}$ converge. Therefore, we adopt a decoupled strategy, freezing $E_{\phi}$ during the initial map optimization to allow $\mathbf{f}$ to stabilize first.
After feature convergence, we periodically adapt $E_{\phi}$ by freezing the map and decoder $D_{\theta}$ and minimizing only $\mathcal{L}_{\text{enc}}$:
\begin{equation}
\label{eq:enc_loss}
\mathcal{L}_{\text{enc}} = \mathcal{L}_{\text{L1}}(E_{\phi}(F_{\text{gt}}), F_{\text{render}})
\end{equation}
This trains $E_{\phi}$ to map the high-dimensional $F_{\text{gt}}$ to the map's learned $F_{\text{render}}$, ensuring stable convergence. This adaptive encoder is critical for real-time 3D object localization, as shown in \cref{sec:ablation}.

\subsection{Language-Guided Gaussian Pruning}
\label{sec:pruning}
To maintain map compactness, we periodically prune Gaussians. Conventional pruning in 3DGS-based systems \cite{kerbl20233d, keetha2024splatam, ha2024rgbd, zhu2025loopsplat} relies on geometric heuristics, such as low opacity or large scale. However, these methods struggle to distinguish redundant Gaussians, and aggressive geometric pruning often degrades map quality by removing structurally important primitives.
To overcome this, we introduce a language-guided pruning strategy that augments the geometric criteria with a semantic criterion, identifying redundant Gaussians that are spatially proximate and semantically similar. This check is efficient as it operates directly on the language features $\mathbf{f}$ already stored on each Gaussian. This eliminates the need for additional feature extraction, unlike semantic pruning methods that require separate models or significant overhead \cite{zhang2024lp, hanson2025pup, ye20243d}.

Specifically, our method identifies redundant primitives by evaluating each Gaussian $G_i$ against its local neighborhood. We first find its K-nearest neighbors in 3D space. Then, for each neighbor $G_j$, we evaluate its redundancy based on both its Euclidean distance $d(\mathbf{p}_i, \mathbf{p}_j)$ and the cosine similarity of their language features, $\text{sim}(\mathbf{f}_i, \mathbf{f}_j)$. A neighbor $G_j$ is marked as redundant if it satisfies the following condition:
\begin{equation}
\label{eq:pruning_condition} (d(\mathbf{p}_i, \mathbf{p}_j) < \tau_{\text{dist}}) \land (\text{sim}(\mathbf{f}_i, \mathbf{f}_j) > \tau_{\text{sim}})
\end{equation}
where $\tau_{\text{dist}}$ and $\tau_{\text{sim}}$ are predefined distance and similarity thresholds. The final set of Gaussians to be pruned is the union of those identified by our language-based redundancy check and those filtered by the traditional geometric criteria.

\subsection{Language-Based Loop Closure}
\label{sec:loop_closure}
To correct the inevitable drift accumulated during long-term operation, we incorporate a language-based loop closure module. Our approach is designed for high efficiency by reusing the features already computed for mapping, avoiding separate loop detection models \cite{mur2017orb, arandjelovic2016netvlad}.

We perform place recognition by representing each keyframe with a compact semantic signature. We utilize a language codebook, generated offline via k-means clustering on high-dimensional language features. This vocabulary allows us to create a normalized histogram for each new keyframe by assigning its language features to the closest clusters in the codebook. We then compute the cosine similarity between the current histogram and those of spatially proximate past keyframes, retaining those above a threshold $\tau_{\text{hist}}$ as loop candidates.

Each loop candidate undergoes a geometric verification using G-ICP against a local submap. If a match is confirmed based on overlap and RMSE thresholds, its relative pose transformation $T_{ij}$ is added as a constraint to a pose graph. We then employ GTSAM \cite{dellaert2012factor} to optimize the entire trajectory for global consistency, propagating the corrected poses back to our 3D Gaussian map.

\section{Experiments}
\label{sec:experiments}

\subsection{Experimental Setup}

\subsubsection{Datasets and Metrics}
We evaluate our framework on the synthetic Replica dataset \cite{straub2019replica} and the real-world TUM-RGBD \cite{sturm2012benchmark} and ScanNet \cite{dai2017scannet} datasets. For tables that report a single metric per dataset, the value represents the average performance across all evaluated scenes. Tracking accuracy is evaluated using Absolute Trajectory Error (ATE RMSE [cm]). Mapping quality is assessed via PSNR, SSIM, and LPIPS. For open-vocabulary performance, we measure semantic understanding using mean Intersection-over-Union (mIoU) and pixel-wise Accuracy, following the protocol of Feature 3DGS \cite{zhou2024feature}.

\subsubsection{Implementation Details}
All experiments are conducted on a desktop equipped with a Ryzen 7 7800X3D CPU, 32GB RAM, and an NVIDIA RTX 4090 GPU. Our system parallelizes tracking and mapping as independent processes via shared memory. Language embeddings are obtained from a CLIP-aligned vision-language extractor~\cite{xie2024sed} running online within the SLAM loop, so the reported FPS includes language feature extraction and the system sustains real-time operation at 15 FPS. The extractor is a modular component of our framework and can be replaced without modifying the rest of the system. Our scene-adaptive encoder and decoder, composed of lightweight 1x1 convolutional layers, are pretrained as an autoencoder to learn the compression to our 16-dimensional space. The language codebook for loop closure is built once offline via k-means clustering on language features from external image datasets~\cite{zhou2017scene, lin2014microsoft}. During the SLAM process, language-guided pruning is executed every 200 iterations. For loop detection, we select the top two candidates with high cosine similarity for geometric verification. To reflect true online performance, all evaluations of SLAM systems are conducted on the final map produced during the SLAM process without post-run optimization, so every SLAM method is evaluated on the live map state available at runtime.

\subsubsection{Baseline}
We evaluate LEGO-SLAM against baselines, with all methods run on the same hardware unless otherwise noted. For open-vocabulary performance, we compare against LeRF \cite{kerr2023lerf}, LangSplat \cite{qin2024langsplat}, and Feature 3DGS \cite{zhou2024feature}. For real-time open-vocabulary mapping, we compare against OLS~\cite{katragadda2025online}, using its published results. For core SLAM capabilities, we benchmark against NeRF-based systems (NICE-SLAM \cite{zhu2022nice}, Point-SLAM \cite{sandstrom2023point}, Loopy-SLAM \cite{liso2024loopy}) and 3DGS-based systems (MonoGS \cite{matsuki2024gaussian}, SplaTAM \cite{keetha2024splatam}, LoopSplat \cite{zhu2025loopsplat}), including those that incorporate loop closure mechanisms.

%%%%%%%%%%%%%%%%%%%%%%%%%%%%%%%%%%%%%%%%%%%%%%%%
\subsection{Quantitative Evaluation}

\subsubsection{Comparison with Real-Time Open-Vocabulary GS-SLAM}
We first compare against Online Language Splatting (OLS)~\cite{katragadda2025online}, the most closely related real-time, open-vocabulary 3DGS mapping system. We compare on Replica, where OLS reports its complete results. To enable a direct comparison, we evaluate under the LangSplat-style text-query protocol that OLS adopts, instead of the Feature 3DGS protocol used in the rest of our experiments. As shown in \cref{tab:ols_comparison}, LEGO-SLAM outperforms OLS in rendering quality (PSNR $+1.2$\,dB) and semantic accuracy (mIoU $+0.056$) while running about $12\times$ faster.
\begin{table}[t]
\centering
\caption{Comparison with Online Language Splatting (OLS)~\cite{katragadda2025online} on Replica (8-scene average). OLS values are paper-reported. Best in \textbf{bold}.}
\label{tab:ols_comparison}
\setlength{\tabcolsep}{4pt}
\renewcommand{\arraystretch}{0.9}
\resizebox{\columnwidth}{!}{%
\begin{tabular}{@{}lcccccc@{}}
\toprule
Method & PSNR$\uparrow$ & SSIM$\uparrow$ & LPIPS$\downarrow$ & mIoU$\uparrow$ & Acc$\uparrow$ & FPS$\uparrow$ \\
\midrule
OLS~\cite{katragadda2025online} & 35.81 & 0.950 & 0.072 & 0.487 & 0.826 & 1.25 \\
LEGO-SLAM & \textbf{36.99} & \textbf{0.965} & \textbf{0.063} & \textbf{0.543} & \textbf{0.827} & \textbf{15.0} \\
\bottomrule
\end{tabular}%
}
\end{table}

\subsubsection{Open-Vocabulary Performance}
\begin{table}[b]
\centering
\caption{Open-Vocabulary Segmentation on 3 datasets. Feature 3DGS failed on several scenes due to memory constraints.}
\label{tab:open_eval_datasets}
    \renewcommand{\arraystretch}{0.9}
    \begin{tabular}{@{}llccc@{}}
    \toprule
    Method & Metrics & Replica & TUM-RGBD & ScanNet \\
    \midrule
    \multirow{2}{*}{LeRF \cite{kerr2023lerf}}
    & Accuracy $\uparrow$ & 0.617 & 0.490 & 0.261 \\
    & mIoU $\uparrow$ & 0.276 & 0.263 & 0.066 \\
    \cmidrule(lr){1-5}
    \multirow{2}{*}{LangSplat \cite{qin2024langsplat}}
    & Accuracy $\uparrow$ & 0.614 & 0.544 & 0.429 \\
    & mIoU $\uparrow$ & 0.263 & 0.229 & 0.160 \\
    \cmidrule(lr){1-5}
    \multirow{2}{*}{Feature 3DGS \cite{zhou2024feature}}
    & Accuracy $\uparrow$ & 0.902 & 0.835 & fail \\
    & mIoU $\uparrow$ & 0.691 & 0.633 & fail \\
    \cmidrule(lr){1-5}
    \multirow{2}{*}{LEGO-SLAM}
    & Accuracy $\uparrow$ & 0.882 & 0.834 & 0.791 \\
    & mIoU $\uparrow$ & 0.674 & 0.650 & 0.519 \\
    \bottomrule
    \end{tabular}%
\end{table}
We evaluate the open-vocabulary mapping performance against representative 3D reconstruction methods, with results summarized in \cref{tab:open_eval_datasets}. For a fair comparison with these baselines, we follow the protocol of \cite{zhou2024feature}, using the per-pixel language embeddings extracted by LSeg \cite{li2022language} as the GT features for all methods and training all baselines for 3000 iterations to match the optimization steps of our mapping module. Only this evaluation uses LSeg as the distillation target, while all other results, including the ablations, are measured in the fully online configuration.
Feature 3DGS \cite{zhou2024feature} achieves high accuracy on Replica by distilling high-dimensional features but suffers from significant memory overhead. Conversely, LangSplat \cite{qin2024langsplat} compresses features to only 3 dimensions, which substantially reduces memory but also leads to significantly lower semantic accuracy. LEGO-SLAM performs competitively with Feature 3DGS on Replica, surpasses it on TUM-RGBD, and scales to all ScanNet scenes where memory-intensive methods fail. Notably, these results are achieved even though our method learns concurrently from estimated camera poses, a more challenging and realistic setting compared to the ground-truth poses used by all baseline methods.

\subsubsection{Tracking and Mapping Performance}
\begin{figure*}[t]
    \centering
    \includegraphics[width=1.0\textwidth]{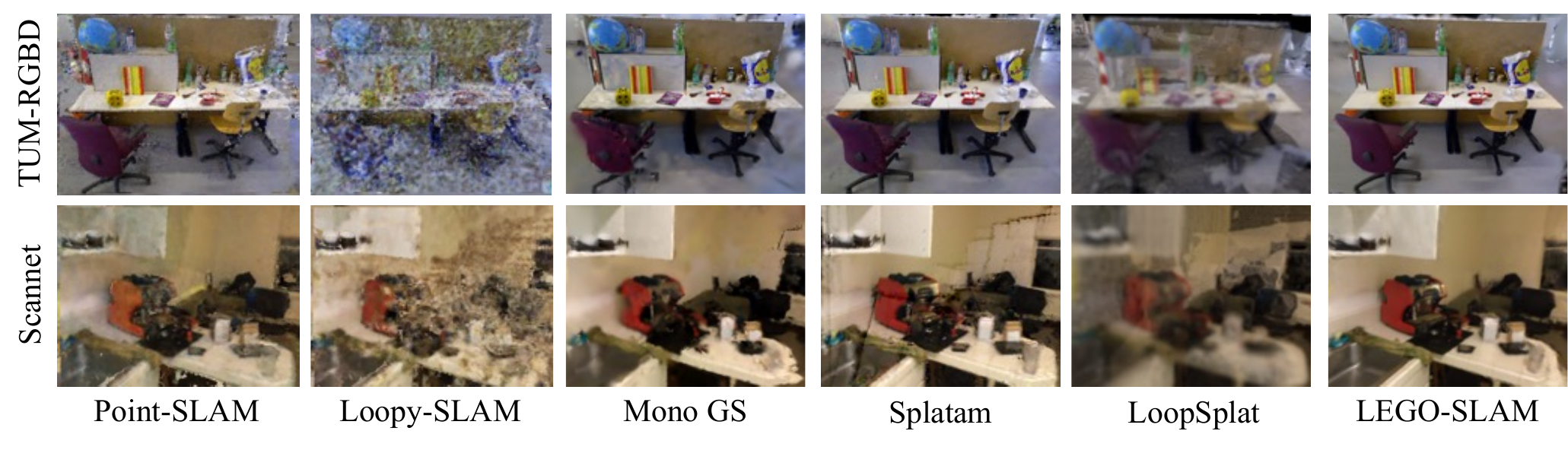}
    \caption{Qualitative Mapping Comparison. Rendered maps of LEGO-SLAM and baselines on TUM-RGBD and ScanNet. All maps are captured from the online SLAM process without post-run optimization.}
    \label{fig:rendering_vis}
\end{figure*}

\begin{table}[!t]
    \centering
    \caption{Tracking Performance on Replica \cite{straub2019replica} (ATE RMSE [cm] $\downarrow$). The best results are highlighted as \colorbox{bestcolor}{\textbf{first}}, \colorbox{secondcolor}{second}, and \colorbox{thirdcolor}{third}.}
    \label{tab:tracking_replica_performance}
    \renewcommand{\arraystretch}{0.9}
    \setlength{\tabcolsep}{4pt}
    \resizebox{\columnwidth}{!}{%
    \begin{tabular}{@{}lccccccccc@{}}
    \toprule
    Method & R0 & R1 & R2 & O0 & O1 & O2 & O3 & O4 & Avg. \\
    \midrule
    NICE-SLAM \cite{zhu2022nice}    & 0.97 & 1.31 & 1.07 & 0.88 & 1.00 & 1.06 & 1.10 & 1.13 & 1.06 \\
    Point-SLAM \cite{sandstrom2023point}    & 0.54 & 0.41 & \cellcolor{thirdcolor}0.23 & \cellcolor{thirdcolor}0.32 & 0.45 & 0.48 & 0.56 & 0.68 & 0.47 \\
    Loopy-SLAM \cite{liso2024loopy}    & \cellcolor{thirdcolor}0.30 & 0.47 & 0.30 & \cellcolor{secondcolor}0.25 & \cellcolor{secondcolor}0.21 & 0.31 & 0.32 & \cellcolor{thirdcolor}0.40 & 0.32 \\
    \midrule
    MonoGS \cite{matsuki2024gaussian}      & \cellcolor{thirdcolor}0.30 & \cellcolor{secondcolor}0.22 & 0.28 & 0.36 & \cellcolor{secondcolor}0.21 & \cellcolor{bestcolor}\textbf{0.24} & \cellcolor{bestcolor}\textbf{0.12} & 0.77 & \cellcolor{thirdcolor}0.31 \\
    SplaTAM \cite{keetha2024splatam}      & 0.31 & 0.39 & 0.27 & 0.49 & \cellcolor{thirdcolor}0.23 & \cellcolor{thirdcolor}0.30 & 0.32 & 0.60 & 0.36 \\
    LoopSplat \cite{zhu2025loopsplat}   & \cellcolor{secondcolor}0.27 & \cellcolor{thirdcolor}0.24 & \cellcolor{bestcolor}\textbf{0.16} & \cellcolor{bestcolor}\textbf{0.23} & \cellcolor{bestcolor}\textbf{0.17} & 0.36 & \cellcolor{secondcolor}0.21 & \cellcolor{secondcolor}0.36 & \cellcolor{secondcolor}0.25 \\
    LEGO-SLAM                 & \cellcolor{bestcolor}\textbf{0.16} & \cellcolor{bestcolor}\textbf{0.18} & \cellcolor{secondcolor}0.18 & \cellcolor{bestcolor}\textbf{0.23} & \cellcolor{thirdcolor}0.23 & \cellcolor{secondcolor}0.25 & \cellcolor{thirdcolor}0.29 & \cellcolor{bestcolor}\textbf{0.24} & \cellcolor{bestcolor}\textbf{0.22} \\
    \bottomrule
    \end{tabular}%
    }

    \vspace{6pt}

    \caption{Tracking Performance on TUM-RGBD \cite{sturm2012benchmark}.}
    \label{tab:tracking_tum_performance}
    \renewcommand{\arraystretch}{0.9}
    \setlength{\tabcolsep}{4pt}
    \begin{tabular*}{\columnwidth}{@{\extracolsep{\fill}}lcccc@{}}
    \toprule
    Method & fr1-desk & fr2-xyz & fr3-office & Avg. \\
    \midrule
    NICE-SLAM \cite{zhu2022nice}    & 2.80 & 2.10 & 7.20 & 4.00 \\
    Point-SLAM \cite{sandstrom2023point}    & 2.73 & \cellcolor{bestcolor}\textbf{1.30} & 3.51 & \cellcolor{thirdcolor}2.51 \\
    Loopy-SLAM \cite{liso2024loopy}    & 3.74 & 1.90 & \cellcolor{thirdcolor}3.12 & 2.92 \\
    \midrule
    MonoGS \cite{matsuki2024gaussian}       & \cellcolor{bestcolor}\textbf{1.47} & 1.57 & \cellcolor{bestcolor}\textbf{1.51} & \cellcolor{bestcolor}\textbf{1.51} \\
    SplaTAM \cite{keetha2024splatam}        & 3.31 & \cellcolor{secondcolor}1.35 & 5.13 & 3.26 \\
    LoopSplat \cite{zhu2025loopsplat}    & \cellcolor{secondcolor}2.35 & \cellcolor{thirdcolor}1.41 & 3.90 & 2.55 \\
    LEGO-SLAM                   & \cellcolor{thirdcolor}2.52 & 1.81 & \cellcolor{secondcolor}2.36 & \cellcolor{secondcolor}2.23 \\
    \bottomrule
    \end{tabular*}

    \vspace{6pt}

    \caption{Tracking Performance on ScanNet \cite{dai2017scannet}.}
    \label{tab:tracking_scannet_performance}
    \renewcommand{\arraystretch}{0.9}
    \setlength{\tabcolsep}{4pt}
    \begin{tabular}{@{}lccccccc@{}}
    \toprule
    Method & 00 & 59 & 106 & 169 & 181 & 207 & Avg. \\
    \midrule
    NICE-SLAM \cite{zhu2022nice}    & 12.0 & 14.0 & \cellcolor{secondcolor}7.9 & 10.9 & 13.4 & 6.2 & 10.73 \\
    Point-SLAM \cite{sandstrom2023point}    & 9.60 & \cellcolor{secondcolor}7.24 & 8.44 & 20.92 & 14.41 & \cellcolor{bestcolor}\textbf{4.56} & 10.86 \\
    Loopy-SLAM \cite{liso2024loopy}    & \cellcolor{bestcolor}\textbf{4.63} & \cellcolor{thirdcolor}7.82 & 8.55  & \cellcolor{secondcolor}7.97 & \cellcolor{secondcolor}11.65 & 6.59 & \cellcolor{secondcolor}7.87 \\
    \midrule
    MonoGS \cite{matsuki2024gaussian}       & 30.34 & 17.03 & 11.30 & 21.53 & 20.51 & 8.07 & 18.13 \\
    SplaTAM \cite{keetha2024splatam}       & 12.19 & 10.01 & 18.05 & 12.37 & \cellcolor{thirdcolor}12.78 & 7.74 & 12.19 \\
    LoopSplat \cite{zhu2025loopsplat}       & \cellcolor{secondcolor}5.30 & \cellcolor{bestcolor}\textbf{7.09} & \cellcolor{bestcolor}\textbf{6.54} & \cellcolor{thirdcolor}10.76 & \cellcolor{bestcolor}\textbf{7.93} & \cellcolor{secondcolor}6.07 & \cellcolor{bestcolor}\textbf{7.28} \\
    LEGO-SLAM                   & \cellcolor{thirdcolor}5.76 & 12.96 & \cellcolor{thirdcolor}8.38 & \cellcolor{bestcolor}\textbf{6.95} & 12.88 & \cellcolor{thirdcolor}6.12 & \cellcolor{thirdcolor}8.84 \\
    \bottomrule
    \end{tabular}%
\end{table}
\begin{table}[!t]
    \centering
    \caption{Rendering Performance and FPS on 3 datasets. All metrics are measured during the online SLAM process without post-run optimization, so baseline scores reflect their live state rather than offline-refined results.}
    \label{tab:rendering_performance}
    \renewcommand{\arraystretch}{0.9}
    \setlength{\tabcolsep}{4pt}
    \begin{tabular}{@{}llccc@{}}
    \toprule
    Method & Metrics & Replica & TUM-RGBD & ScanNet \\
    \midrule
    \multirow{4}{*}{Point-SLAM \cite{sandstrom2023point}}
    & PSNR[dB] $\uparrow$ & \cellcolor{secondcolor}35.56 & \cellcolor{thirdcolor}21.33 & \cellcolor{bestcolor}\textbf{23.31} \\
    & SSIM $\uparrow$      & \cellcolor{bestcolor}\textbf{0.977} & \cellcolor{thirdcolor}0.733 & \cellcolor{secondcolor}0.753 \\
    & LPIPS $\downarrow$   & \cellcolor{thirdcolor}0.118 & 0.453 & \cellcolor{thirdcolor}0.509 \\
    & FPS $\uparrow$       & 0.415 & 0.252 & 0.233 \\
    \midrule
    \multirow{4}{*}{Loopy-SLAM \cite{liso2024loopy}}
    & PSNR[dB] $\uparrow$ & 19.28 & 14.32 & 12.46 \\
    & SSIM $\uparrow$      & 0.662 & 0.512 & 0.495 \\
    & LPIPS $\downarrow$   & 0.506 & 0.470 & 0.574 \\
    & FPS $\uparrow$       & 0.374 & 0.222 & 0.357 \\
    \midrule
    \multirow{4}{*}{MonoGS \cite{matsuki2024gaussian}}
    & PSNR[dB] $\uparrow$ & \cellcolor{thirdcolor}35.33 & 17.82 & 16.23 \\
    & SSIM $\uparrow$      & 0.943 & 0.714 & 0.599 \\
    & LPIPS $\downarrow$   & 0.122 & \cellcolor{thirdcolor}0.327 & 0.588 \\
    & FPS $\uparrow$       & \cellcolor{secondcolor}0.679 & \cellcolor{secondcolor}2.52 & \cellcolor{secondcolor}2.01 \\
    \midrule
    \multirow{4}{*}{SplaTAM \cite{keetha2024splatam}}
    & PSNR[dB] $\uparrow$ & 34.19 & \cellcolor{secondcolor}23.53 & \cellcolor{thirdcolor}18.82 \\
    & SSIM $\uparrow$      & \cellcolor{secondcolor}0.970 & \cellcolor{bestcolor}\textbf{0.908} & \cellcolor{thirdcolor}0.699 \\
    & LPIPS $\downarrow$   & \cellcolor{secondcolor}0.094 & \cellcolor{secondcolor}0.166 & \cellcolor{secondcolor}0.370 \\
    & FPS $\uparrow$       & 0.212 & 0.407 & \cellcolor{thirdcolor}0.544 \\
    \midrule
    \multirow{4}{*}{LoopSplat \cite{zhu2025loopsplat}}
    & PSNR[dB] $\uparrow$ & 14.13 & 12.85 & 14.20 \\
    & SSIM $\uparrow$      & 0.748 & 0.511 & 0.571 \\
    & LPIPS $\downarrow$   & 0.584 & 0.746 & 0.708 \\
    & FPS $\uparrow$       & \cellcolor{thirdcolor}0.651 & \cellcolor{thirdcolor}0.58 & 0.445 \\
    \midrule
    \multirow{4}{*}{LEGO-SLAM}
    & PSNR[dB] $\uparrow$ & \cellcolor{bestcolor}\textbf{36.99} & \cellcolor{bestcolor}\textbf{24.72} & \cellcolor{secondcolor}21.15 \\
    & SSIM $\uparrow$      & \cellcolor{thirdcolor}0.965 & \cellcolor{secondcolor}0.861 & \cellcolor{bestcolor}\textbf{0.795} \\
    & LPIPS $\downarrow$   & \cellcolor{bestcolor}\textbf{0.063} & \cellcolor{bestcolor}\textbf{0.131} & \cellcolor{bestcolor}\textbf{0.257} \\
    & FPS $\uparrow$       & \cellcolor{bestcolor}\textbf{15.0} & \cellcolor{bestcolor}\textbf{15.0} & \cellcolor{bestcolor}\textbf{15.0} \\
    \bottomrule
    \end{tabular}%
\end{table}
We evaluate the tracking accuracy against representative NeRF-based and 3DGS-based SLAM systems (\cref{tab:tracking_replica_performance,tab:tracking_tum_performance,tab:tracking_scannet_performance}).
On the Replica dataset, LEGO-SLAM achieves the lowest average ATE (0.22 cm). It remains competitive on the challenging real-world TUM-RGBD sequences (2.23 cm) and large-scale ScanNet (8.84 cm). This ScanNet performance, which leverages our efficient language-based loop closure, approaches specialized loop-closure systems like LoopSplat and Loopy-SLAM while running over 30$\times$ faster, and exceeds the accuracy of all non-loop-closure baselines. For mapping quality (\cref{tab:rendering_performance}), our method is again the best or highly competitive across all datasets, and \cref{fig:rendering_vis} shows the corresponding qualitative comparison. All these results are obtained at 15 FPS, while no competing system exceeds 3 FPS. Notably, this performance is achieved while concurrently building a rich, open-vocabulary map, whereas the baselines only perform geometric and RGB mapping.

% %%%%%%%%%%%%%%%%%%%%%%%%%%%%%%%%%%%%%%%%

\begin{table}[t!]
\centering
\caption{Ablation on Initialization. Convergence speed in iterations comparing pretrained initialization against training from scratch, where convergence is defined as Total loss $<$ 0.1, RGB $<$ 0.05, and Feature $<$ 0.05.}
\label{tab:convergence_comparison}
    \renewcommand{\arraystretch}{0.9}
    \setlength{\tabcolsep}{4pt}
    \begin{tabular}{@{}llccc@{}}
    \toprule
    Method & Conv. Steps & Replica & TUM-RGBD & ScanNet \\
    \midrule
    \multirow{3}{*}{Ours (w/o Init)}
    & RGB           & \textbf{217} & \textbf{235} & 298 \\
    & Feature       & 229 & 233 & 236 \\
    & Total         & 220 & 232 & 254 \\
    \cmidrule(lr){1-5}
    \multirow{3}{*}{Ours (w/ Init)}
    & RGB           & 218 & 242 & \textbf{267} \\
    & Feature       & \textbf{79} & \textbf{64} & \textbf{63} \\
    & Total         & \textbf{180} & \textbf{151} & \textbf{148} \\
    \bottomrule
    \end{tabular}%

    \vspace{8pt}

    \includegraphics[width=\linewidth]{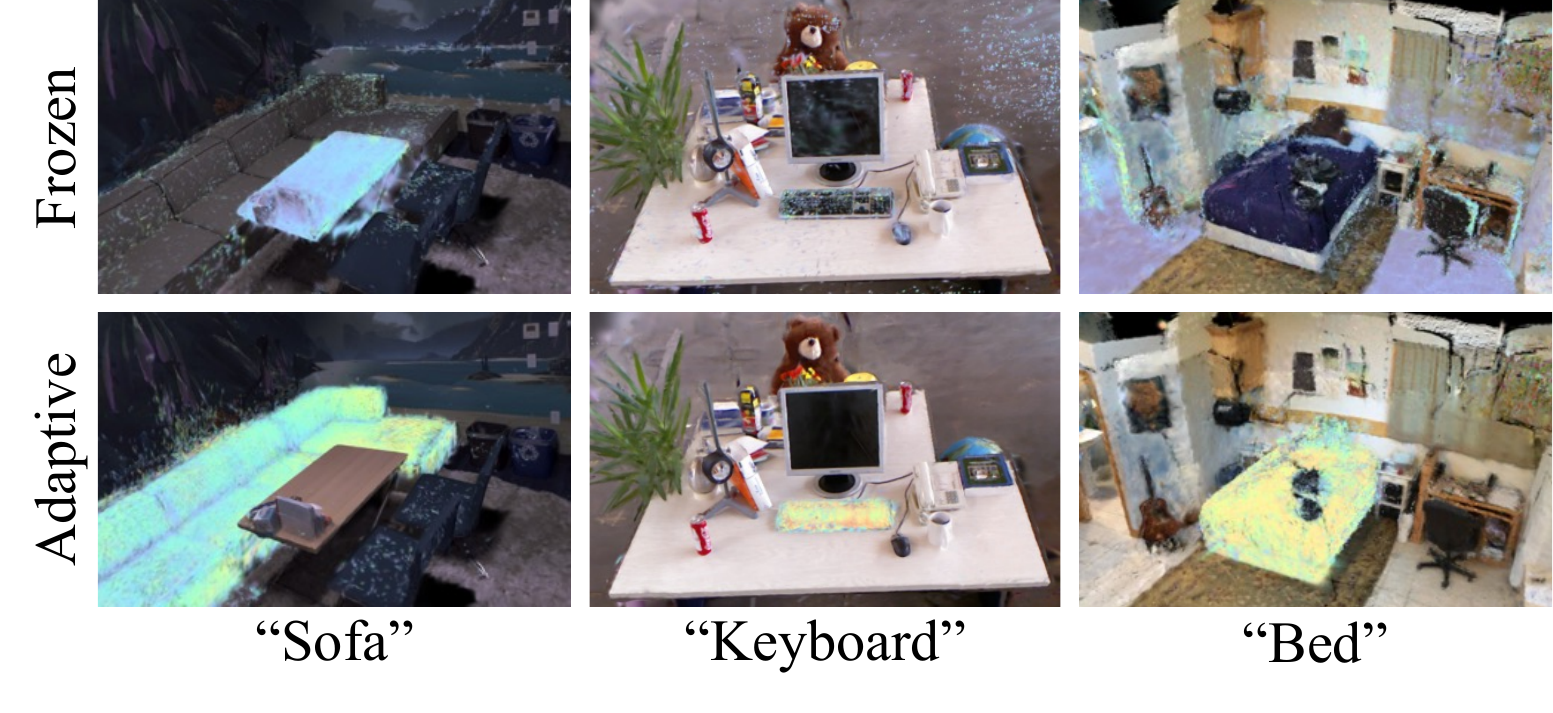}
    \captionof{figure}{Qualitative Analysis of Encoder Adaptation. Our scene-adaptive encoder accurately localizes 3D objects by continuously learning scene-specific features online, whereas the frozen baseline fails to capture semantic context.}
    \label{fig:adaptive_vis}
\end{table}

\subsection{Ablation Study}
\label{sec:ablation}

\subsubsection{Analysis of the Scene-Adaptive Encoder}
We validate our encoder-based initialization strategy. As shown in \cref{tab:convergence_comparison}, training from scratch (Ours w/o Init) requires substantially more iterations for the feature loss to converge. In contrast, using our pretrained encoder as a strong prior for new Gaussians (Ours w/ Init) considerably accelerates convergence.

While pretraining provides a strong prior, continuous adaptation remains critical in SLAM, where the system must learn from new environments. To demonstrate this, we compare our adaptive encoder against a frozen pretrained encoder. As visualized in \cref{fig:adaptive_vis}, the frozen encoder fails to produce meaningful localization results, as its static features cannot adapt to the specific scene, whereas our scene-adaptive encoder successfully localizes the object. This shows that continuous online learning is essential for our open-vocabulary capability.

\begin{table}[!t]
\centering
\caption{Impact of feature dimension ($d$).}
\label{tab:feature_dim_ablation}
    \renewcommand{\arraystretch}{0.9}
    \setlength{\tabcolsep}{4pt}
    \begin{tabular}{@{}llccc@{}}
    \toprule
    Method & Metrics & Replica & TUM-RGBD & ScanNet \\
    \midrule
    \multirow{3}{*}{Ours (d=8)}
    & PSNR [dB] $\uparrow$   & \textbf{37.29} & \textbf{24.93} & \textbf{21.22} \\
    & Accuracy $\uparrow$    & 0.789 & 0.790 & 0.648 \\
    & Memory [MB] $\downarrow$ & \textbf{61.9} & \textbf{71.5} & \textbf{194.6} \\
    \midrule
    \multirow{3}{*}{Ours (d=16)}
    & PSNR [dB] $\uparrow$   & 36.99 & 24.72 & 21.15 \\
    & Accuracy $\uparrow$    & \textbf{0.822} & \textbf{0.838} & \textbf{0.700} \\
    & Memory [MB] $\downarrow$ & 81.8 & 92.9 & 256.6 \\
    \midrule
    \multirow{3}{*}{Ours (d=32)}
    & PSNR [dB] $\uparrow$   & 36.46 & 24.24 & 20.91 \\
    & Accuracy $\uparrow$    & 0.791 & 0.815 & 0.673 \\
    & Memory [MB] $\downarrow$ & 123.7 & 146.1 & 385.6 \\
    \midrule
    \multirow{3}{*}{Ours (d=64)}
    & PSNR [dB] $\uparrow$   & 35.57 & 23.84 & -- \\
    & Accuracy $\uparrow$    & 0.788 & 0.812 & -- \\
    & Memory [MB] $\downarrow$ & 201.3 & 248.1 & -- \\
    \midrule
    \multirow{3}{*}{Ours (d=128)}
    & PSNR [dB] $\uparrow$   & 32.58 & -- & -- \\
    & Accuracy $\uparrow$    & 0.712 & -- & -- \\
    & Memory [MB] $\downarrow$ & 371.1 & -- & -- \\
    \bottomrule
    \end{tabular}%
\end{table}

\begin{figure*}[t]
    \centering
    \includegraphics[width=1.0\textwidth]{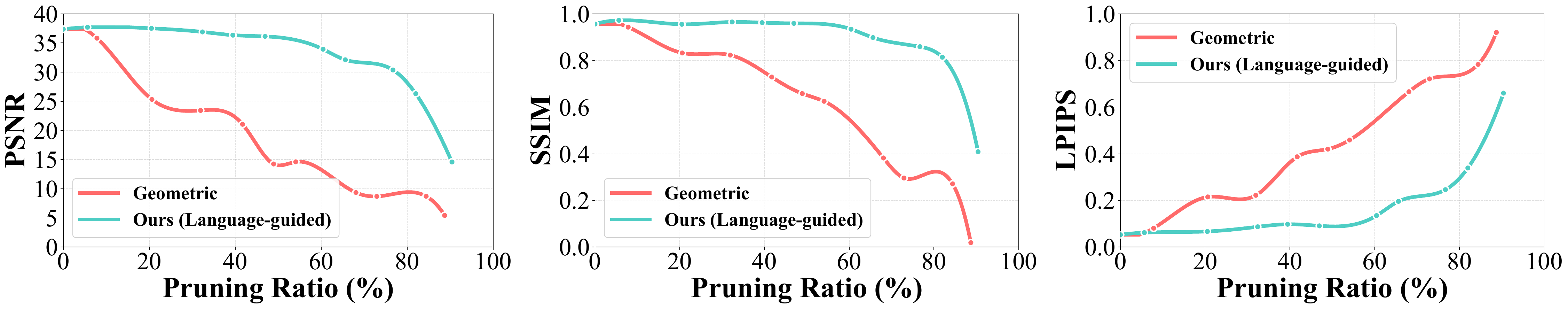}
    \caption{Pruning Performance Comparison. As the pruning ratio increases, our language-guided method shows significantly less degradation in rendering quality compared to the geometric approach on the Replica dataset.}
    \label{fig:prun_graph}
\end{figure*}

\subsubsection{Impact of Feature Dimension}
We investigate the impact of the feature dimension ($d$) on mapping quality, semantic accuracy, and memory usage in \cref{tab:feature_dim_ablation}. A larger $d$ increases memory and rendering costs, which in turn reduces the mapping optimization iterations available in real-time SLAM. At $d=8$, the system is most efficient, yielding the highest PSNR as geometry converges more fully, but its semantic accuracy is markedly lower than at $d=16$, as 8 dimensions are insufficient for feature reconstruction. Conversely, at $d \ge 32$, semantic accuracy also drops because slower rendering reduces mapping iterations, preventing feature convergence. At $d=64$, the memory overhead becomes prohibitive on the large-scale ScanNet scenes, and at $d=128$ it fails even on TUM-RGBD. Based on this analysis, we select $d=16$, accepting a slight PSNR drop relative to $d=8$ in exchange for a significant gain in semantic accuracy.

\subsubsection{Analysis of Language-Guided Pruning}
\begin{table}[!t]
    \centering
    \caption{Pruning Ablation Study. We compare our language-guided
    strategy against a geometric-only baseline under a highly aggressive
    setting. Bold marks the better of the two pruning strategies.}
    \label{tab:pruning_ablation_study}
    \renewcommand{\arraystretch}{0.9}
    \setlength{\tabcolsep}{4pt}
    \begin{tabular}{@{}clccc@{}}
    \toprule
    Method & Metrics & Replica & TUM-RGBD & ScanNet \\
    \midrule
    \multirow{6}{*}{\begin{tabular}{@{}c@{}}Baseline \\ (No Pruning)\end{tabular}}
    & PSNR $\uparrow$      & 37.35 & 24.39 & 20.28 \\
    & SSIM $\uparrow$      & 0.962 & 0.858 & 0.797 \\
    & LPIPS $\downarrow$   & 0.053 & 0.141 & 0.257 \\
    & Acc $\uparrow$       & 0.819 & 0.841 & 0.712 \\
    & IoU $\uparrow$       & 0.597 & 0.689 & 0.454 \\
    & \# of GS (K) $\downarrow$ & 621 & 789 & 1796 \\
    \midrule
    \multirow{6}{*}{\begin{tabular}{@{}c@{}}Baseline \cite{keetha2024splatam, zhu2025loopsplat} \\ (Geo Pruning)\end{tabular}}
    & PSNR $\uparrow$      & 25.32 & 17.87 & 17.65 \\
    & SSIM $\uparrow$      & 0.832 & 0.731 & 0.707 \\
    & LPIPS $\downarrow$   & 0.215 & 0.261 & 0.357 \\
    & Acc $\uparrow$       & 0.774 & 0.761 & 0.632 \\
    & IoU $\uparrow$       & 0.520 & 0.627 & 0.387 \\
    & \# of GS (K) $\downarrow$ & 493 & 623 & 1657 \\
    \cdashline{1-5}
    \addlinespace
    \multirow{6}{*}{\begin{tabular}{@{}c@{}}\textbf{Ours} \\ \textbf{(Lang Pruning)}\end{tabular}}
    & PSNR $\uparrow$      & \textbf{36.34} & \textbf{22.88} & \textbf{20.17} \\
    & SSIM $\uparrow$      & \textbf{0.956} & \textbf{0.818} & \textbf{0.776} \\
    & LPIPS $\downarrow$   & \textbf{0.098} & \textbf{0.176} & \textbf{0.281} \\
    & Acc $\uparrow$       & \textbf{0.811} & \textbf{0.818} & \textbf{0.680} \\
    & IoU $\uparrow$       & \textbf{0.587} & \textbf{0.681} & \textbf{0.431} \\
    & \# of GS (K) $\downarrow$ & \textbf{376} & \textbf{334} & \textbf{1370} \\
    \bottomrule
    \end{tabular}%
\end{table}
We analyze the effectiveness of our language-guided pruning against the conventional geometric pruning adopted in \cite{keetha2024splatam, zhu2025loopsplat}, comparing their performance as the pruning ratio increases. As visualized in \cref{fig:prun_graph}, the rendering quality of geometric pruning degrades sharply even at low pruning ratios, whereas our language-guided method maintains high quality across a much wider range. This is confirmed by \cref{tab:pruning_ablation_study}, where under a highly aggressive setting, our method preserves PSNR and semantic accuracy while removing even more Gaussians than the geometric baseline, reducing the map by up to 58\% relative to the unpruned map. This is achieved with negligible overhead, as the semantic check reuses the 16-dimensional features already stored on the Gaussians.

\subsubsection{Effectiveness of Language-Based Loop Detection}
We validate our language-based loop closure, which reuses the language features already computed during mapping and thus avoids the overhead of separate detection models \cite{mur2017orb, arandjelovic2016netvlad}. To demonstrate that this efficiency does not sacrifice accuracy, we compare against a similarly lightweight, position-based baseline \cite{shan2020lio}. As shown in \cref{tab:loop_detection_comparison}, our method consistently achieves lower tracking error across all datasets, indicating more robust place recognition.

\begin{table}[!t]
\centering
\caption{Loop Detection Comparison. Our language-based method achieves lower tracking error (ATE RMSE [cm] $\downarrow$) than the position-based approach.}
\label{tab:loop_detection_comparison}
    \setlength{\tabcolsep}{4pt}
    \begin{tabular}{@{}lccc@{}}
    \toprule
    Method & Replica & TUM-RGBD & ScanNet \\
    \midrule
    Position-based \cite{shan2020lio} & 0.28 & 3.13 & 10.19 \\
    Language-based  & \textbf{0.22} & \textbf{2.23} & \textbf{8.84} \\
    \bottomrule
    \end{tabular}
\end{table}

\section{Conclusion}
\label{sec:conclusion}
We proposed LEGO-SLAM, a real-time, open-vocabulary 3DGS-based SLAM system. Our core contribution is a lightweight, scene-adaptive autoencoder that distills language features into a compact 16-dimensional space, considerably reducing memory and rendering overhead. These compact features further enable a language-guided pruning strategy, and the language features extracted for mapping are reused for efficient loop closure. Experiments demonstrated that LEGO-SLAM achieves competitive mapping quality and tracking accuracy at 15 FPS, with language feature extraction running online within the SLAM loop.

%%%%%%%%%%%%%%%%%%%%%%%%%%%%%%%%%%%%%%%%%%%%%%%%%%%%%%%%%%%%%%%%%%%%%%%%%%%%%%%%

\bibliographystyle{IEEEtran}
\bibliography{main}

\end{document}